\documentclass[sigconf]{aamas}

\usepackage{balance} 

\usepackage{algorithm}
\usepackage{algorithmic}
\usepackage{amsmath}
\usepackage{amsthm}
\usepackage{cite}
\usepackage{tikz}
\usepackage{xcolor}
\usepackage{multirow}
\usepackage{makecell}
\usepackage{listings}
\usepackage{tabularx}

\newcommand{\newtext}[1]{{#1}}

\doi{CBJO8409}

\makeatletter
\gdef\@copyrightpermission{
  \begin{minipage}{0.2\columnwidth}
   \href{https://creativecommons.org/licenses/by/4.0/}{\includegraphics[width=0.90\textwidth]{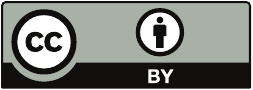}}
  \end{minipage}\hfill
  \begin{minipage}{0.8\columnwidth}
   \href{https://creativecommons.org/licenses/by/4.0/}{This work is licensed under a Creative Commons Attribution International 4.0 License.}
  \end{minipage}
  \vspace{5pt}
}
\makeatother

\setcopyright{ifaamas}
\acmConference[AAMAS '26]{Proc.\@ of the 25th International Conference
on Autonomous Agents and Multiagent Systems (AAMAS 2026)}{May 25 -- 29, 2026}
{Paphos, Cyprus}{C.~Amato, L.~Dennis, V.~Mascardi, J.~Thangarajah (eds.)}
\copyrightyear{2026}
\acmYear{2026}
\acmDOI{}
\acmPrice{}
\acmISBN{}

\title[RUMAD]{RUMAD: Reinforcement-Unifying Multi-Agent Debate}

\subtitle{AAAI Track}

\author{Chao Wang}
\affiliation{
  \institution{Tsinghua University}
  \city{Shenzhen}
  \country{China}}
\email{wangchao23@mails.tsinghua.edu.cn}

\author{Han Lin}
\affiliation{
  \institution{Zhejiang University}
  \city{Hangzhou}
  \country{China}}
\affiliation{
  \institution{Tsinghua University}
  \city{Shenzhen}
  \country{China}}
\email{hlin@zju.edu.cn}

\author{Huaze Tang}
\affiliation{
  \institution{Tsinghua University}
  \city{Shenzhen}
  \country{China}}
\email{thz21@mails.tsinghua.edu.cn}

\author{Huijing Lin}
\affiliation{
  \institution{Tsinghua University}
  \city{Shenzhen}
  \country{China}}
\email{linhj24@mails.tsinghua.edu.cn}

\author{Wenbo Ding}
\affiliation{
  \institution{Tsinghua University}
  \city{Shenzhen}
  \country{China}}
\email{ding.wenbo@sz.tsinghua.edu.cn}

\begin{abstract}
Multi-agent debate (MAD) systems leverage collective intelligence to enhance reasoning capabilities, yet existing approaches struggle to simultaneously optimize accuracy, consensus formation, and computational efficiency. Static topology methods lack adaptability to task complexity variations, while external LLM-based coordination risks introducing privileged knowledge that compromises debate neutrality. This work presents \textbf{RUMAD (Reinforcement-Unifying Multi-Agent Debate)}, a novel framework that formulates dynamic communication topology control in MAD as a reinforcement learning (RL) problem.

RUMAD employs a content-agnostic observation scheme that captures high-level debate dynamics avoiding access to raw agent reasoning content. RUMAD uses a multi-objective reward to model solution quality, cohesion and efficiency. A PPO-trained controller dynamically adjusts edge weights in the communication graph, while a dual-threshold mechanism enables fine-grained control over both agent activation and information visibility.

Experimental evaluation across MMLU, GSM8K, and GPQA benchmarks demonstrates that RUMAD achieves substantial efficiency gains—reducing token costs by over 80\%—while still improving reasoning accuracy compared to single LLM model and multiple MAD baselines. Notably, RUMAD trained exclusively on MMLU exhibits robust zero-shot generalization to out-of-domain (OOD) tasks, indicating that the learned communication strategies capture task-independent principles of effective multi-agent coordination. These results establish RUMAD as a efficient and robust approach for deploying  multi-agent reasoning application with practical resource constraints.
\end{abstract}

\keywords{Multi-agent Debate, LLM, Reinforcement Learning}

\newcommand{\BibTeX}{\rm B\kern-.05em{\sc i\kern-.025em b}\kern-.08em\TeX}

\begin{document}


\pagestyle{fancy}
\fancyhead{}


\maketitle 


\section{Introduction}

\begin{figure}[t]
  \includegraphics[width=\columnwidth]{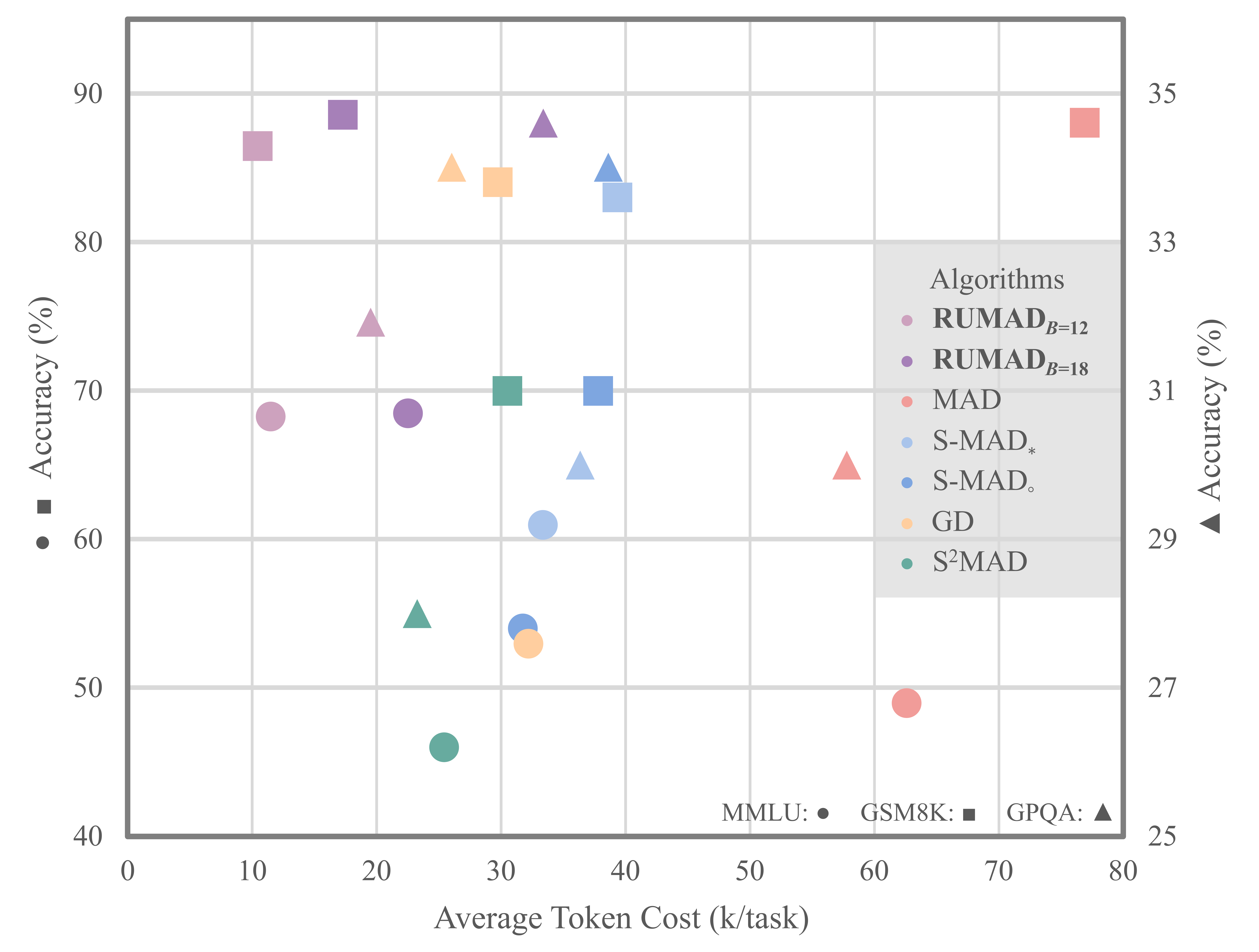}
  \caption{\newtext{Comparison of accuracy and average token cost. 
  Accuracy for MMLU ($\bullet$) and GSM8K ($\blacksquare$) corresponds to the left Y-axis, while GPQA (\textbf{$\blacktriangle$}) corresponds to the right Y-axis. 
  Points closer to the top-left corner represent a superior accuracy-efficiency trade-off. 
  The results show that RUMAD variants consistently outperform baselines by achieving higher accuracy at a lower token cost.}}
  \label{fig:RUMAD_comparison}
  \Description{Comparison of accuracy and token consumption. The results show that RUMAD achieves better trade-off between accuracy and efficiency.}
\end{figure}

\begin{table*}[t]
\centering
\caption{Different types of MAD methods are designed from different strategies. Mark $\bullet$ indicates corresponding metrics are taken into design target. Mark $\circ$ indicates methods may have side-effect on corresponding metrics but don't focus on them. Mark - means methods fail on corresponding aspects.}
\begin{tabular}{lccc}
\toprule
                            & Static Topology Methods & External-LLM Methods & \textbf{RUMAD}  \\ \midrule
Accuracy                    & $\bullet$                  & $\bullet$               & $\bullet$ \\
Consensus                   & $\circ$                    & $\bullet$               & $\bullet$ \\
External Opinion Independent & $\bullet$                  & -                    & $\bullet$ \\
Question Independent        & $\bullet$                  & -                    & $\bullet$ \\
Communication Cost Control  & $\circ$                    & $\circ$                 & $\bullet$ \\ \bottomrule
\end{tabular}
\label{tab:Methods_list}
\end{table*}

Multi-agent debate (MAD) frameworks harness the collective intelligence of multiple large language models (LLMs) \citep{touvron2023llama,zhao2023survey, naveed2023comprehensive,jiang2024survey,achiam2023gpt,glm2024chatglm,guo2025deepseek} to achieve superior reasoning and decision-making beyond what any single model can attain. However, the effectiveness and practicality of MAD systems fundamentally depend on how agents communicate and exchange information during debate \citep{liang2023encouraging}.

\newtext{\textbf{Adaptivity}:} Most existing MAD approaches utilize static sparse topologies, such as Sparse MAD (S-MAD) \citep{li2024improving}, Group Debate (GD) \citep{liu2024groupdebate} or Selective Sparse MAD (S$^2$-MAD) \citep{zeng2025s} using ring, star, or fixed grouped/hierarchical structures, to reduce the communication burden compared to fully connected networks \citep{du2023improving,hill2015real,sun2023corex}. Such methods  inherently lack flexibility. \newtext{Fixed topologies cannot adapt to the evolving diversity, difficulty, or semantic style changes present across different tasks (e.g., allocating the same budget to an easy and a hard problem).} As a result, they easily either cut off valuable information paths for complex tasks or retain redundant, low-value connections on simple tasks, ultimately hampering both efficiency and collective reasoning. 

\newtext{\textbf{Privacy}: Some adaptive methods introduce external LLM-based “judger” or “summarizer” roles, where a larger often more powerful agent (e.g GPT-4) supervises the debate. 
Context within a debate is exposed to external models. External knowledge in return suppresses the diversity and emergent collaboration of internal agent swarm.}

\newtext{\textbf{Efficiency}:}A further critical limitation of existing MAD approaches is the lack of explicit modeling of token cost. Most previous MAD designs focus solely on accuracy or consensus, overlooking the real-world computational and financial constraints imposed by large-scale LLM deployments. Without a principled framework for managing communication budgets, these systems either waste excessive resources or struggle to generalize under practical constraints. Static methods typically impose a hard upper bound on communication rounds by restricting the debate topology. 

Major challenges arise in balancing accuracy, diversity and efficiency(communication budget) of MAD, as show in Table~\ref{tab:Methods_list}. \newtext{To address these challenges, we propose \textbf{RUMAD} (\textbf{\textit{Reinforcement-Unifying Multi-Agent Debate}}), a novel framework that formulates dynamic agent coordination and topology control as a reinforcement learning (RL) problem, without external LLM utilization.} 

RUMAD formulates dynamic communication topology control as a reinforcement learning (RL) problem that is fundamentally independent of debate content. Specifically, we represent the communication topologies among agents as a dynamic weighted directed graph, where a non-LLM RL agent dynamically adjusts the edge weights in response to the evolving debate context. \newtext{Crucially, the RL controller is \textbf{content-agnostic}}: it does not intervene in the reasoning process or rely on privileged knowledge, but instead observes \newtext{\textbf{high-level structural dynamics}} such as semantic similarity \newtext{scalars}, answer agreement, debate progress, and communication cost. By modeling observations in this manner, RUMAD preserves the neutrality of dynamic debate management and effectively mitigates the risk of introducing external opinions or privileged information into the reasoning process. 

RUMAD constructs a multi-objective reward function that explicitly balances accuracy, consensus, and token efficiency, while strictly enforcing communication budget constraints. We employ the PPO algorithm to train the topology control agent, which can achieve these trade-offs after only lightweight training. More importantly, the content-agnostic and neutral design of RUMAD endows it with strong zero-shot generalization capability, enabling robust performance across diverse tasks and domains without the need for task-specific finetuning.

Our experiments on standard benchmarks, including \textbf{MMLU} \citep{hendrycks2020measuring}, \textbf{GSM8K} \citep{cobbe2021training}, and \textbf{GPQA} \citep{reinGPQAGraduateLevelGoogleProof2023}, demonstrate that RUMAD achieves a superior balance between collective accuracy and token usage compared to both fully-connected and static sparse baselines. \newtext{As shown in Figure~\ref{fig:RUMAD_comparison}, take GSM8K for example, RUMAD$_{B=12}$ requires only 0.121k tokens per percentage point of accuracy, and RUMAD$_{B=18}$ requires 0.195k, compared to the 0.354k tokens required by the second-best baseline (GD).} Concretely, RUMAD reduces token cost by 80\% while improving accuracy on MMLU, and robustly generalizes to new tasks (GPQA, GSM8K) with similar performance.

\textbf{Our contributions are summarized as follows:}

\begin{itemize}
    \item We propose RUMAD, a novel reinforcement learning framework that dynamically controls multi-agent debate topologies to optimize both accuracy and efficiency.
    \item We design a unified, content-agnostic observation and reward scheme that enables robust and generalizable learning of communication strategies across tasks and agent configurations.
    \item We empirically demonstrate that RUMAD outperforms both fully connected and static sparse MAD baselines, achieving superior trade-offs between performance and cost on multiple challenging benchmarks.
\end{itemize}

\begin{figure*}[t]
  \includegraphics[width=\textwidth]{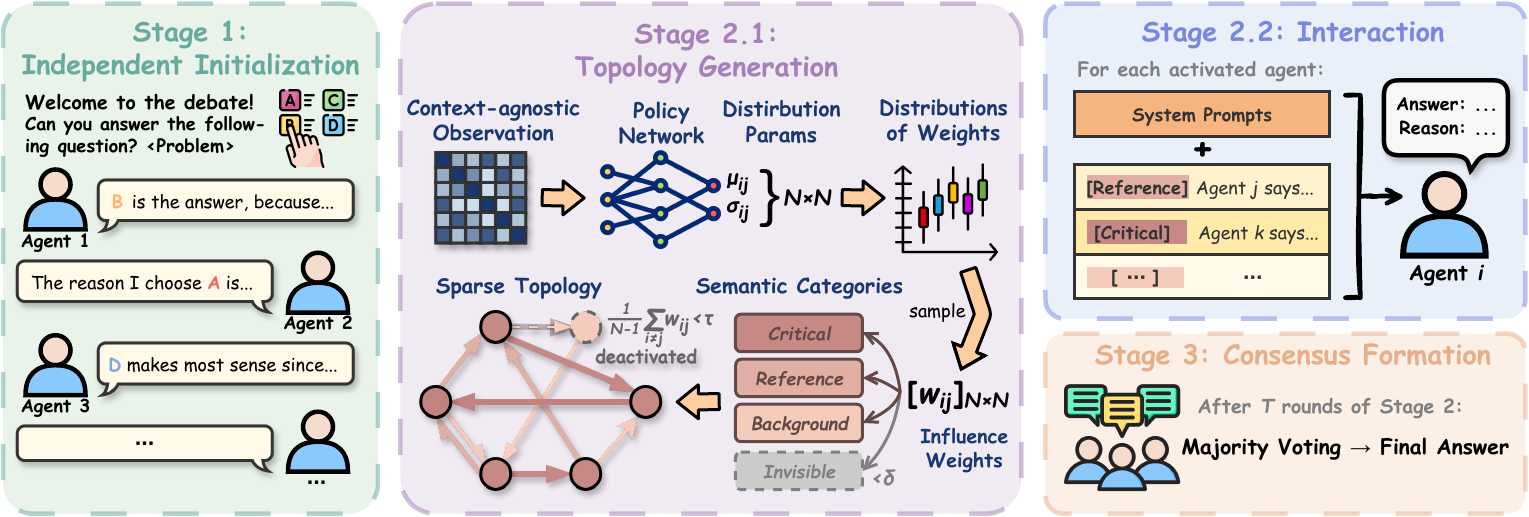}
  \caption{The process pipeline of RUMAD. In the first stage, all agents gives the initial response. In the second stage, RUMAD organizes topology in stage 2.1 and the selected agents debate with each other in stage 2.2. In the last stage, the final decision is obtained via majority voting.}
  \Description{The process pipeline of RUMAD. In the first stage, all agents gives the initial response. In the second stage, RUMAD organizes topology in stage 2.1 and the selected agents debate with each other in stage 2.2. In the last stage, the final decision is obtained via majority voting.} 
  \label{fig:process_pipeline}
\end{figure*}

\section{Preliminaries}
\subsection{Multi-Agent Debate}
Multi-agent debate (MAD) is an emergent paradigm for collective reasoning, where multiple language model agents iteratively exchange their intermediate thoughts and conclusions in order to reach a more accurate and robust consensus than any single agent alone. A central research focus within MAD is the design of communication topologies—that is, how agents share information and interact across debate rounds.

Early studies typically employed fully connected topologies, allowing all agents to access each other's responses at every round. However, such dense communication can be inefficient and costly, particularly as the number of agents grows. To address this, subsequent works have explored various forms of sparse and structured topologies. For instance, S-MAD \citep{li2024improving} adopts fixed sparse patterns such as rings (only adjacent agents are visible) or stars (only centric agent are visible), enabling each agent to communicate only with a subset of neighbors, thereby reducing token consumption. Other approaches, like Group Debate (GD) \citep{liu2024groupdebate}, partition agents into static groups, which debate internally before group-level aggregation. Similarly, two-stage topologies such as S$^2$-MAD \citep{zeng2025s} first organize local debates within small groups and then aggregate results through a secondary global round if group opinions differ. These strategies all aim to balance the trade-off consensus formation, and computational efficiency.

\subsection{Static Workflow Generation and Dynamic Topology Control}

\newtext{Multi-agent debate (MAD) systems rely on efficient inter-agent communication topologies. This challenge shares many similarities with recent research about automatic generation of agent workflows. such as MACNET \citep{qian2025ScalingLargeLanguage}, AFlow \citep{zhang2025AFlowAutomatingAgentic}, and MaAS \citep{zhang2025MultiagentArchitectureSearch}, which primarily focus on optimizing the static workflow or agentic pipeline. For example, MACNET analyzes the performance of predefined static topologies (e.g., chain, mesh) . AFlow uses MCTS to search for a single, optimal static workflow for a given task domain. MaAS refines this by using a controller to select a single, query-dependent static workflow at the beginning of a task. Despite above advances, however, most existing multi-agent LLM frameworks rely on static or manually specified topologies, which cannot adapt to the dynamic needs or varying complexity of different tasks.}

This motivates the development of adaptive and principled methods for controlling agent interactions. \newtext{Reinforcement learning (RL) is an end-to-end solution which is capable of both modeling and scheduling temporal dynamics during debate. Hierarchical RL frameworks like FMH \citep{ahilan2019FeudalMultiAgentHierarchies} use multi-faceted rewards, where the objective is semantic task decomposition via manager-worker subgoaling.}

\newtext{Our work addresses a fundamentally different problem. RUMAD is not concerned with generating a static, task-level workflow. Instead, we focus on dynamic, round-by-round communication management within a debate. Our RL controller is invoked at each step to adaptively reconfigure the communication graph based on the evolving state of the debate (e.g., agent agreement, semantic similarity). This allows RUMAD to achieve fine-grained efficiency, such as pruning connections to agents that have already reached consensus, a capability not present in static workflow models. Considering hierarchical reward modeling, RUMAD's multi-objective reward function is novel in its content-agnostic nature, balancing the system-level trade-offs between solution quality, group cohesion, and computational cost.}

\subsection{Markov Decision Process}
A Markov Decision Process (MDP) is a mathematical framework for modeling sequential decision-making problems, where an agent interacts with an environment over discrete time steps. 
At each step, the agent observes the current state, selects an action, and receives a reward and a new state from the environment. The objective is to find a policy that maximizes the expected cumulative reward \citep{suttonReinforcementLearningIntroduction2018}.
In our multi-agent debate (MAD) setting, this framework allows the controller to base its topology decisions solely on the observable debate state at each round, ensuring a neutral and adaptive decision process that does not require access to agents’ internal semantic content.

\subsection{Proximal Policy Optimization}
Proximal Policy Optimization (PPO) \citep{schulmanProximalPolicyOptimization2017a} is a widely used reinforcement learning algorithm for training policies in high-dimensional and dynamic environments. PPO employs an actor-critic framework, where a policy network proposes actions and a value network estimates the expected return for each state. By optimizing a clipped surrogate objective, PPO achieves stable and efficient policy updates, and has demonstrated robust performance across various sequential decision-making tasks.
In our MAD framework, the value network provides reliable estimates of the debate state’s long-term potential, which in turn stabilizes and guides the learning of RUMAD about adaptive topology control strategies. Additionally, we design a multi-objective reward model to capture the diverse goals of the debate process, and employ Generalized Advantage Estimation (GAE) \citep{schulmanHighDimensionalContinuousControl2018} to further improve learning efficiency and convergence.

\section{Method}

We propose a reinforcement learning (RL) framework for adaptive topology control in multi-agent debate (MAD), in which a centralized controller dynamically regulates inter-agent communication to maximize collective reasoning performance under communication constraints. Our approach is characterized by the following innovations: (1) a content-agnostic observation scheme that leverages only interaction-relevant features without accessing agent-level semantics, (2) a multi-objective reward formulation balancing accuracy, consensus, efficiency, and sparsity, (3) a soft and differentiable action space modeling each edge's communication weight as a stochastic variable, and (4) a dual mechanism for communication efficiency combining budget constraints and agent pruning. The full pipeline consists of three stages: initial independent agent reasoning, iterative topology-controlled sparse debate, and consensus aggregation via majority voting.
\subsection{Framework Pipeline}
\label{sec:framework_pipeline}
To illustrate the operation process of RUMAD, we decompose our framework into three sequential stages, each responsible for
distinct aspects of multi-agent debate, as show in Figure~\ref{fig:process_pipeline}.

\paragraph{\textbf{Stage 1: Independent Initialization}}
All agents start by independently generating their initial responses to the given problem. Each agent $i$ produces an initial response $r_i^0$ and conclusion $c_i^0$ based solely on the input prompt, without knowledge of other agents’ perspectives, ensuring diverse hypotheses that span the solution space.

\paragraph{\textbf{Stage 2: Adaptive Sparse Debate}}
In this stage, agents iteratively interact under dynamically controlled topologies. We divide the process into two focused steps:

\begin{enumerate}
\item \emph{Topology Generation:} The PPO policy network ingests the current observations and outputs the influence weights $w_{ij}^t$ that determines how agents exchange information. Agents with insufficient incoming influence will be masked and reuse their previous responses to save tokens. Thus, we obtain the active nodes and the directed edges with different weights, which together constitute the interaction topology.
\item \emph{Interaction:} For each active agent $i$, the debate environment construct customized prompts by grouping neighbors into categories based on weight magnitude (Critical, Reference, Background, Invisible). This interaction step propagates information along pruned links, enabling focused debate under a communication budget.
\end{enumerate}

\paragraph{\textbf{Stage 3: Consensus Formation}}
After $T$ adaptive rounds, the final conclusions $\{c_i^T\}$ are aggregated via majority voting. The consensus answer emerges as the label with the highest agent support, reflecting the collective reasoning of the ensemble rather than any single participant.


\subsection{Content-Agnostic Observation}

At each debate round $t$, the controller observes a global state $o_t$ constructed without reference to raw agent content. For each agent $i$, let $r_i^t$ denote its reasoning embedding and $c_i^t$ its current answer. The embedded observable state for actor and critic is identically defined by:
\begin{equation}
o_t = \mathrm{MLP}(\left[ \mathrm{sim}(i,j)\right]_{N\times N})
\end{equation}
where the $N\times N$ similarity matrix is given \newtext{by MAD environment}:
\begin{equation}
\mathrm{sim}(i, j) = \lambda \cos(r_i^t, r_j^t) + (1-\lambda)\, \mathbb{I}(c_i^t = c_j^t)
\label{eq:similarity}
\end{equation}
with hyperparameter $\lambda \in [0,1]$, cosine similarity $\cos(\cdot, \cdot)$ between reasoning embeddings, and $\mathbb{I}(\cdot)$ the indicator function for answer agreement. \newtext{The observation thus encodes only answer-level agreement and high-level semantic structure, ensuring content-agnostic control. RUMAD scheduler can only observe an $N \times N$ matrix with floating-point similarity number within $[0,1]$ without accessing the raw text (nor embedding of text) of agent reasoning.}

\subsection{Action Space and Edge Weight Distribution Modeling}

In our framework, the action space at each debate round consists of a set of continuous communication weights, where each possible directed edge $(i, j)$ is assigned a real-valued weight $w_{ij}^t \in (0, 1)$ reflecting the influence of agent $j$ on agent $i$. To enable principled policy optimization via stochastic gradient methods, we adopt a probabilistic action parameterization: the policy network (actor) outputs, for each edge, the parameters $(\mu_{ij}^t, \sigma_{ij}^t)$ of a Gaussian distribution, from which the edge weight is sampled as
\begin{equation}
w_{ij}^t \sim \mathrm{Sigmoid}\big(\mathcal{N}(\mu_{ij}^t,\, {\sigma_{ij}^t}^2)\big).
\label{eq:weight_definition}
\end{equation}
This formulation satisfies the philosophy of PPO. The use of a Gaussian distribution, followed by a Sigmoid transformation to enforce the valid range, provides a flexible and fully differentiable mechanism for modeling continuous actions. This allows for efficient exploration of the topology space, mitigates overfitting to deterministic patterns, and enables robust credit assignment across complex, high-dimensional action spaces. Such stochastic action parameterizations are standard in modern RL for continuous control (e.g., PPO).

\subsection{Communication Budget Constraint}

To achieve joint optimization of reasoning accuracy and communication efficiency, we introduce a dual-threshold mechanism that directly links the learned edge weights to both token consumption and semantic information flow within the debate process. 

Specifically, at each round, the sampled weight matrix $W^t = [w_{ij}^t]_{N \times N}$ not only determines the semantic relevance of each neighbor's response for prompt construction \newtext{(edge-wise)}, but also governs whether an agent actively updates its opinion  \newtext{(node-wise)}, thereby incurring token cost.

\begin{figure}[t]
  \includegraphics[width=\columnwidth]{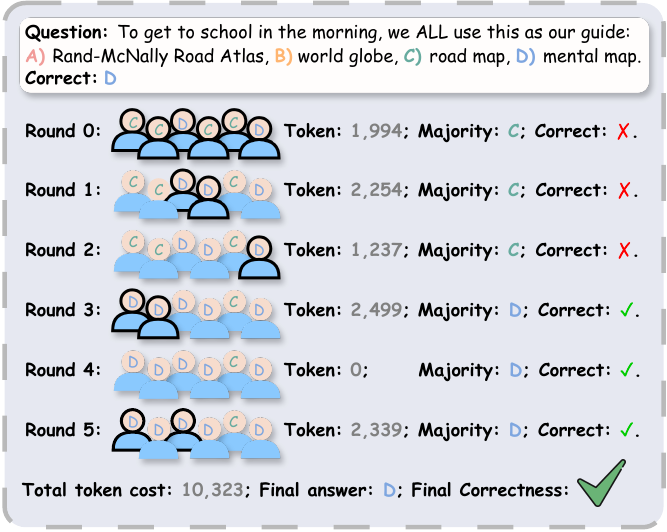}
  \caption{An example debate controlled by RUMAD.}
  \label{fig:RUMAD_example}
  \Description{An example debate controlled by RUMAD.}
\end{figure}

For each agent $i$, we compute its average external influence \newtext{(in-degree)} as
\begin{equation}
\bar{w}_i^t = \frac{1}{N-1} \sum_{j\neq i} w_{ij}^t,
\end{equation}
where $N$ is the total number of agents. An activation threshold $\tau$ is applied: if $\bar{w}_i^t < \tau$, agent $i$ is pruned for the current round and reuses its previous response. \newtext{For simplicity, we utilize $w_{ii}$ as activation threshold $\tau$. The $\tau = w_{ii}^t$ design dynamically models implicit "confidence" of each agent. An agent is confident enough while assigned with higher weight on its own opinion than the average of the external, $\bar{w}_i^t < w_{ii}^t$. As shown by example round 4 in Figure~\ref{fig:RUMAD_example}, it results in zero token usage this round that all agents are assigned as confident. In the sub-sequential round 5, different weights are sampled from distributions defined in Equation~\ref{eq:weight_definition} thus more communication occurs.} This threshold thus directly controls the expected number of agents generating new tokens per round, providing a fine-grained handle on collective communication expenditure.

The computed weights determine not only the influence strength but also how information is presented to each agent. We implement a three-tier prompt priority filter:
\begin{equation}
P_{ij}^t = \begin{cases}
\text{[Critical]} & \text{if } w_{ij}^t > 0.40 \\
\text{[Reference]} & \text{if } w_{ij}^t > 0.25 \\
\text{[Background]} & \text{if } w_{ij}^t > 0.10
\end{cases}
\label{eq:semantic_magnitude}
\end{equation}
Agents prioritize debate context with tags $P_{ij}^t$ quantized from computed influence weights $w_{ij}^t$, while maintaining the natural language interaction paradigm of LLMs.

To regularize overall communication, we introduce a global budget parameter $B$ representing a soft upper bound on the total number of active communication links per round. Formally, the set of active links is defined as $\{ (i,j) : w_{ij}^t > \delta \}$(where $\delta=0.10$ as minimum visible weight shown in Equation~\ref{eq:semantic_magnitude}), and a budget penalty is incorporated into the policy loss whenever this set exceeds $B$:
\begin{equation}
\mathcal{L}_\mathrm{budget}(\theta) = \max\left(0,\, \sum_{i,j} \mathbb{I}(w_{ij}^t > \delta) - B \right),
\label{eq:Loss_budget}
\end{equation}
where $\mathbb{I}(\cdot)$ is the indicator function. \newtext{Crucially, $B$ acts as a \textbf{training-time prior} via the loss function, encouraging the policy to internalize a preference for sparsity, rather than acting as a \textbf{hard constraint at inference}.} This budget constraint reflects a prior preference for overall communication compactness, while leaving the allocation of communication opportunities across agents to be adaptively learned by the policy to best balance informativeness and efficiency.

Through this budget-regularized design, RUMAD enables precise, learnable trade-offs between debate effectiveness and communication cost, granting the controller the capacity to both minimize unnecessary token usage and adaptively maintain critical informational flows.

\subsection{Multi-Objective Reward}
\newtext{To guide the RL agent in navigating the complex trade-offs of debate, we designed a hierarchical, multi-objective reward function. Our design is structured around three core objectives: }

\newtext{\textbf{Maximizing Solution Quality.}}
As major rewarding signal, $\mathrm{Accuracy}_{t/{ep}}$ are 0/1 binary rewards reflecting whether the majority answers match the ground truth. 
$\mathrm{Progress}_t$ applies a round-driven-decaying reward amplification to $\mathrm{Accuracy}_t$ when round answer is correct. Thus correctness in early-stage is encouraged. 
$\mathrm{Improvement}_{t/{ep}}$ are also binary signals rewarding shift from wrong answer to correct answer across adjacent rounds or between round 0 and final response.

\newtext{\textbf{Promoting Group Cohesion.}}
$\mathrm{Consensus}_{t/{ep}}$ measures the proportion of agent answers that match current majority answer, together with their semantic similarity, rewarding more unified and coherent group responses.

\newtext{\textbf{Optimizing Resource Efficiency.}}
$\mathrm{Efficiency}_t$ incentivizes reduced token consumption relative to a baseline answer-length budget. 
In addition, a penalty signal, $\mathrm{Sparsity}_t$, is designed as indicator of "schedule expense". Expenses are assigned corresponding to semantic magnitude filtered in Equation~\ref{eq:semantic_magnitude}.

Formally, at each debate round $t$, the controller receives a composite reward $R_t$ that integrates six key factors:
\begin{align}
R_t = \;& \alpha_1\,\mathrm{Accuracy}_t + \alpha_2\,\mathrm{Consensus}_t + \alpha_3\,\mathrm{Progress}_t \\
 &+ \alpha_4\,\mathrm{Efficiency}_t + \alpha_5\,\mathrm{Improvement}_t - \alpha_6\,\mathrm{Sparsity}_t. \nonumber
\end{align}
where $\mathrm{Progress}_t$ and $\mathrm{Sparsity}_t$ are designed as in-progress-only signals. At the end of each debate episode, we compute a terminal reward $R_{{ep}}$:
\begin{align}
R_{{ep}} =\; & \beta_1\,\mathrm{Accuracy}_{{ep}} + \beta_2\,\mathrm{Consensus}_{{ep}} \\
& + \beta_3\,\mathrm{Efficiency}_{{ep}}  + \beta_4\,\mathrm{Improvement}_{{ep}} \nonumber
\end{align}


\newtext{$R_t$ and $R_{{ep}}$ are roughly homogeneous with different preference. As shown in Table~\ref{tab:reward_weight_parameter}, parameters of $R_t$ and $R_{{ep}}$ are distributed similarly, except $\mathrm{Improvement}_{t}$ are emphasized with higher weight in $R_t$ as encouragement of runtime adjustment. This two-tier reward mechanism enables the controller to balance immediate step-wise gains (e.g., rapid consensus or token savings) with ultimate debate-level objectives, facilitating robust, generalizable policy learning for efficient multi-agent reasoning under budget constraints. Structural necessity of this hierarchical design is supported by ablation in Section~\ref{sec:ablation}. And parameter robustness is shown through 0-shot generalization revealed in Section~\ref{sec:main_result}.}

\subsection{RL Dynamic Topology Control}

The controller is trained using the Proximal Policy Optimization (PPO) framework. The policy loss is given by
\begin{equation}
\mathcal{L}_\pi(\theta) = - \mathbb{E}_t \left[ \min\left( r_t\,\hat{A}_t,\, \mathrm{clip}(r_t, 1-\varepsilon, 1+\varepsilon)\hat{A}_t \right) \right]
\end{equation}
where $r_t = \frac{\pi_\theta(a_t|o_t)}{\pi_{\theta_\mathrm{old}}(a_t|o_t)}$ is the probability ratio, and $\hat{A}_t$ is the advantage computed via Generalized Advantage Estimation (GAE) using the composite reward. 

The PPO value network $\phi$ is trained to minimize the squared error:
\begin{equation}
\mathcal{L}_v(\phi) = \mathbb{E}_t \left[ (V_\phi(o_t) - V_t^\mathrm{target})^2 \right]
\end{equation}
with $V_\phi(o_t)$ as predicted expectation of future return with current observation $o_t$ versus $V_t^\mathrm{target}$ as the bootstrapped return at time step $t$. 

The total loss for joint PPO training is
\begin{equation}
\mathcal{L}(\theta, \phi) = \mathcal{L}_\pi(\theta) + c_1\,\mathcal{L}_v(\phi) + c_2\,\mathcal{L}_\mathrm{budget}(\theta)
\end{equation}
where $c_1, c_2 \geq 0$ \newtext{balance loss signals among policy, value and regularization into comparable orders of magnitude, according to PPO style}.

Training proceeds by iteratively sampling debate episodes, updating the controller with the composite loss, and periodically evaluating performance under fixed communication budgets. At deployment, the controller deterministically selects edge weights using the mean of the learned Gaussian distributions.

\section{Experiments}
\subsection{Experimental Setup}
\label{sec:setup}

We conduct experiments on three challenging and complementary benchmarks: \textbf{MMLU} \citep{hendrycks2020measuring}, \textbf{GSM8K} \citep{cobbe2021training}, and \textbf{GPQA} \citep{reinGPQAGraduateLevelGoogleProof2023}. The PPO-based topology controller of RUMAD is exclusively trained on the MMLU dev dataset, leveraging its broad coverage of multi-domain reasoning tasks. Table~\ref{tab:hyperparameters} summarizes the training hyperparameter settings used throughout our experiments. For evaluation, we report results on the MMLU test set as well as on GSM8K (arithmetic reasoning) and GPQA (graduate-level science QA). Notably, for GSM8K and GPQA, we apply RUMAD controller in a zero-shot transfer setting without any further fine-tuning, directly assessing the generalization capability of the learned communication policy.


\begin{table}[t]
\centering
\caption{Performance of RUMAD and baselines across three evaluation datasets. Token cost is calculated as average of each topic debated. (6,6) is general mark of MAD systems indicating 6-agent, 6-round configuration. RUMAD significantly reduces token cost with comparable or improved accuracy. S-MAD with different structures are denoted as * (Star) and o (Ring). Since RUMAD is trained upon MMLU dev partition, MMLU test partition is used for evaluation. RUMAD evaluated upon GPQA-main and GSM8K is zero-shot migrated, without any finetuning with new datasets. RUMAD shows superior cross-domain generality. \protect\newtext{Note: All baseline accuracies are measured under identical 4-bit quantization and 0-shot settings for a fair comparison.}} 
\begin{tabular}{llccc}
\toprule
\textbf{Task} & \textbf{Method} (6,6) & \textbf{ACC} & \textbf{\makecell[c]{Token Cost\\(k/task)}} & \textbf{\makecell[c]{Cost\\Saving}} \\
\midrule
\multirow{7}{*}{MMLU}
 & \textbf{RUMAD$_{B=12}$} & 68\% & \textbf{11.43} & \textbf{-81.74\%} \\
 & \textbf{RUMAD$_{B=18}$} & \textbf{68\%} & 22.46 & -64.11\% \\
 & MAD & 49\% & 62.58 & N/A \\
 & S-MAD$_*$ & 61\% & 33.32 & -46.75\% \\
 & S-MAD$_\circ$ & 54\% & 31.70 & -49.34\% \\
 & GD & 53\% & 32.17 & -48.59\% \\
 & S$^2$MAD & 46\% & 25.36 & -58.48\% \\
\midrule
\multirow{7}{*}{GPQA}
 & \textbf{RUMAD$_{B=12}$} & 32\% & \textbf{19.53} & \textbf{-66.21\%} \\
 & \textbf{RUMAD$_{B=18}$} & \textbf{35\%} & 33.39 & -42.24\% \\
 & MAD & 30\% & 57.80 & N/A \\
 & S-MAD$_*$ & 30\% & 36.37 & -37.08\% \\
 & S-MAD$_\circ$ & 34\% & 38.59 & -33.24\% \\
 & GD & 34\% & 26.05 & -54.93\% \\
 & S$^2$MAD & 28\% & 23.26 & -59.76\% \\
\midrule
\multirow{7}{*}{GSM8K}
 & \textbf{RUMAD$_{B=12}$} & 86\% & \textbf{10.46} & \textbf{-86.40\%} \\
 & \textbf{RUMAD$_{B=18}$} & \textbf{89\%} & 17.28 & -77.53\% \\
 & MAD & 88\% & 76.90 & N/A \\
 & S-MAD$_*$ & 83\% & 39.38 & -48.78\% \\
 & S-MAD$_\circ$ & 70\% & 37.82 & -50.81\% \\
 & GD & 84\% & 29.76 & -61.30\% \\
 & S$^2$MAD & 70\% & 30.51 & -60.32\% \\
\bottomrule
\end{tabular}
\label{tab:performance}
\end{table}

To maximize the heterogeneity and dynamic interactions within the debate, we instantiate a diverse agent pool comprising three different LLM models: \textbf{LLaMA-3.1-8B-Instruct} \citep{touvron2023llama}, \textbf{ChatGLM-4-9B} \citep{glm2024chatglm}, and \textbf{Deepseek-Math-7B-Instruct} \citep{guo2025deepseek}. Each architecture is represented by two agents, forming a six-member debate group per trial. \newtext{For each agent, there are 4 non-similar agents and 1 similar agent.} This multi-architecture composition is designed to capture a spectrum of reasoning styles and knowledge bases. 

For computational efficiency and scalability, all LLMs are deployed using \newtext{\textbf{4-bit groupwise quantization (Q4\_K\_M)}} with mixed precision. Embedding model used for RUMAD to compute similarity among texts, defined in Equation~\ref{eq:similarity}, is \textbf{Nomic-Embed-v1} \citep{nussbaumNomicEmbedTraining2025}. This configuration enables concurrent inference of all agents on a single NVIDIA GeForce RTX 3090 GPU, facilitating systematic ablation and large-scale evaluation. \newtext{Thus, an efficiency-focused \textbf{0-shot, 4-bit} setup is crucial for fair baseline comparison. For instance, the LLaMA-3.1-8B-Instruct is officially reported with 69.4\% accuracy on MMLU under \textbf{5-shot, bf16-precision} conditions, while actually achieving 57.9\% under \textbf{0-shot, 4-bit} setting.} 

We benchmark RUMAD primarily against MAD, a fully-connected debate protocol that omits both visibility control and prompt structuring, thereby preserving complete information exchange among all agents throughout the discussion. In addition, we evaluate against S-MAD, GD, and S$^2$-MAD using each method’s reported optimal configuration. 

Our evaluation focuses on two key metrics: (1) solution \textbf{accuracy} across diverse problem categories, and (2) computational \textbf{efficiency} as measured by total token consumption. And we evaluate generality of RUMAD by comparing its performance across different benchmarks. Scripts for training and evaluating, along with RUMAD PPO checkpoint weight files, will be provided in supplementary files.

\subsection{Results and Analysis}
\label{sec:main_result}
As shown in Table~\ref{tab:performance}, our proposed RUMAD framework achieves a substantial reduction in communication cost while maintaining, or even enhancing, answer accuracy across all evaluated benchmarks. On both MMLU and GSM8K, RUMAD reduces average token usage per task by over 80\% compared to the fully connected MAD baseline, yet achieves much better accuracy (e.g., 68\% vs 49\% on MMLU). RUMAD also provides best token cost saving on GPQA, while higher accuracy is available if assigned more communication budget (12 to 18 times per round). This demonstrates the efficacy of reinforcement learning-based dynamic topology control of RUMAD in identifying and preserving the most informative agent interactions under tight communication budgets. The results highlight that significant efficiency gains are attainable without sacrificing group reasoning performance.

\paragraph{\textbf{$B$ as configurable prior budget}} A salient advantage of our framework lies in the explicit budget parameter $B$ utilized in budget loss defined in Equation~\ref{eq:Loss_budget}, which directly governs the maximum expected communication load per round. By adjusting $B$, practitioners can specify a clear prior on resource consumption, aligning debate efficiency with practical deployment constraints. Our results show that even at relatively low budget levels (e.g., $B=12$ for six agents), the controller learns to allocate communication judiciously, maintaining high accuracy with a fraction of the original token cost.

We configured a comparison with $B=18$, indicating user prior preference is communication of 3 times per agent per round in identical 6-agent system. RUMAD is capable to allocate this budget adaptively and get higher accuracy performance on all benchmarks, with corresponding trade-off on token-cost. This interpretability and controllability of $B$ enhances the usability of our approach in real-world multi-agent systems. As we can see, on MMLU tasks, $B=12$ budget is sufficient for RUMAD to coordinate debate and achieve comparable accuracy with $B=18$ budget (68.30\% vs 68.48\%), while on GPQA and GSM8K, RUMAD utilize more budget to provide better performance.

\newtext{As defined in Equation~\ref{eq:Loss_budget}, $B$ acts as a \textbf{training-time prior} via the loss function, encouraging the policy to internalize a preference for sparsity, rather than a \textbf{hard constraint at inference} applied to scheduler. Utilization of $B$ provides better alignment of human prior preference without increasing cost of hyperparameter tuning.}

\paragraph{\textbf{RUMAD zero-shot generality}} We observe that the RUMAD controller, trained solely on the MMLU development set, generalizes robustly to out-of-domain benchmarks such as GPQA and GSM8K in a zero-shot setting. It's notable that MMLU is single-choice task while GSM8K needs arithmetic answers. Despite no exposure to these domains during training, RUMAD maintains competitive accuracy and cost savings, underscoring the generality and transferability of the learned communication policy. \newtext{This strong out-of-domain performance also serves as key evidence of \textbf{hyperparameter robustness}. The same model trained on MMLU under reward model with parameter$\alpha_k, \beta_k$ transferred directly to GSM8K and GPQA without modification, indicating the learned policy is not brittle or overfitted to the training domain, a common concern with complex reward structures.} This result suggests that our method is not tightly coupled to any single task distribution or agent type, facilitating practical deployment across diverse collaborative reasoning scenarios.

\begin{table}[t]
\setlength{\tabcolsep}{1.3mm}
\centering
\caption{\newtext{Ablation study of RUMAD's key components. We report accuracy (ACC, \%) and average token cost (TC,  in k/task). The best result for each metric is in \textbf{bold}. We set $B=12$ for all variants.}}
\begin{tabular}{lcccccc}
\toprule
\multirow{2.5}{*}{\textbf{Ablation Module}} & \multicolumn{2}{c}{\textbf{MMLU}} & \multicolumn{2}{c}{\textbf{GPQA}} & \multicolumn{2}{c}{\textbf{GSM8K}} \\
\cmidrule(lr){2-3} \cmidrule(lr){4-5} \cmidrule(lr){6-7}
 & \textbf{ACC} & \textbf{TC} & \textbf{ACC} & \textbf{TC} & \textbf{ACC} & \textbf{TC} \\
\midrule
\textbf{RUMAD} & 68.3 & \textbf{11.4} & \textbf{31.9} & \textbf{19.5} & \textbf{86.4} & \textbf{10.5} \\
$\beta$ distribution & 65.1 & 18.7 & 29.2 & 28.7 & 84.5 & 17.2 \\
w/o $R_{ep}$ & 66.9 & 15.6 & 31.0 & 24.4 & 85.1 & 14.3 \\
w/o $R_{t}$ & 67.2 & 22.2 & 30.1 & 32.6 & 86.0 & 20.2 \\
w/o agent activation & 65.4 & 51.9 & 31.7 & 67.5 & 86.1 & 45.0 \\
w/o $\mathcal{L}_\mathrm{budget}$ & \textbf{70.0} & 17.4 & 27.7 & 26.8 & 85.8 & 15.6 \\
\bottomrule
\end{tabular}
\label{tab:ablation}
\end{table}

\subsection{\newtext{Ablation Study}}
\label{sec:ablation}
\newtext{We conduct a comprehensive ablation study to investigate the contribution of each key component of RUMAD. To ensure a fair comparison, all variants are trained exclusively on the MMLU dev dataset and operate under a consistent 6-agent, 6-round configuration. The results, summarized in Table~\ref{tab:ablation}, are evaluated on MMLU, GPQA, and GSM8K, reporting both accuracy and average token cost. The findings demonstrate that each module is integral to the model's performance, and the full RUMAD framework achieves the best overall trade-off between accuracy and efficiency.}

\newtext{\paragraph{\textbf{Change to $\beta$ distribution}} RUMAD models agent connections using a Gaussian distribution to avoid overfitting. Replacing this with a $\beta$ distribution results in a significant performance drop: accuracy falls across all benchmarks (e.g., 68.3\% to 65.1\% on MMLU), and the token cost substantially increases, underscoring the superior modeling capability of the Gaussian parameterization.}

\newtext{\paragraph{\textbf{Ablation of $R_{ep}$ and $R_t$}} Our two-tier reward system is critical for balancing immediate and long-term goals. It combines the per-round reward $R_{t}$ with an episode reward $R_{ep}$ applied only at the final round. In thew/o $R_{ep}$ variant, we replace the final round's $R_{ep}$ with the standard $R_{t}$, which causes a notable drop in accuracy and a $\sim$37\% increase in token cost on MMLU (11.4k to 15.6k). Conversely, in the w/o $R_{t}$ variant, we broadcast the final $R_{ep}$ value to all preceding rounds. This results in a severe impact on efficiency, nearly doubling the token cost. This highlights that both reward signals are necessary for learning an efficient policy.}

\newtext{\paragraph{\textbf{Ablation of agent activation}} The agent activation mechanism is paramount for efficiency. It conditionally decides if an agent generates a new response, enabling true communication sparsification. Removing it (w/o agent activation) means that even with efficient pruning of communication edges, all agents are still forced to generate a new response in every round. This inability to dynamically silence agents causes a catastrophic increase in token cost (e.g., a $\sim$4.5x increase on MMLU), while accuracy also suffers. This confirms that dynamic agent activation, not just edge pruning, is the primary driver of RUMAD's token savings.}

\newtext{\paragraph{\textbf{Ablation of $\mathcal{L}_\mathrm{budget}$}} The budget loss $\mathcal{L}_\mathrm{budget}$ acts as a crucial regularizer for both efficiency and generalization. Without it, the controller fails to converge to an efficient policy, incurring higher token costs. Interestingly, while this variant achieves higher accuracy on the in-domain MMLU task (70.0\%), its performance drops significantly on the out-of-domain benchmarks GPQA (27.7\% vs. 31.9\%) and GSM8K (85.8\% vs. 86.4\%). This strongly suggests that $\mathcal{L}_\mathrm{budget}$ not only enforces efficiency but also guides the RL algorithm to learn more generalizable communication strategies.}

\begin{table}[t]
\centering
\caption{\newtext{Scalability analysis with varying agent numbers. We report Accuracy (ACC) and average Token Cost (TC, k/task) on MMLU and GSM8K.}}
\label{tab:scaling}
\begin{tabular}{cccccc}
\toprule
\multirow{2.5}{*}{\textbf{Agents} ($N$)} & \multirow{2.5}{*}{\textbf{Budget} ($B$)} & \multicolumn{2}{c}{\textbf{MMLU}} & \multicolumn{2}{c}{\textbf{GSM8K}} \\
\cmidrule(lr){3-4} \cmidrule(lr){5-6}
 & & \textbf{ACC} & \textbf{TC} & \textbf{ACC} & \textbf{TC} \\
\midrule
3 & 6 & 66.6\% & 2.3 & 85.0\% & 2.0 \\
6 & 12 & 68.3\% & 11.4 & 86.4\% & 10.5 \\
6 & 18 & 68.5\% & 22.5 & \textbf{88.6\%} & 17.3 \\
9 & 18 & \textbf{70.3\%} & 17.6 & 88.5\% & 15.9 \\
\bottomrule
\end{tabular}
\end{table}

\subsection{\newtext{Scalability Analysis}}

\newtext{We investigate RUMAD's scalability and robustness by evaluating its performance across varying agent group sizes ($N \in \{3, 6, 9\}$) and budgets ($B \in \{6, 12, 18\}$), with results summarized in Table~\ref{tab:scaling}. For a fair comparison, all hyperparameters except $B$ (e.g., learning rate, reward coefficients) are kept identical, demonstrating the model's robustness.
}

\newtext{First, the results demonstrate a clear positive correlation between the number of agents and reasoning performance. When scaling the agent count $N$ and communication budget $B$ proportionally ($N \in \{3, 6, 9\}$ while $B=2 \times N$), we observe consistent improvements in accuracy—rising from 66.6\% to 70.3\% on MMLU and from 85.0\% to 88.5\% on GSM8K. This trend indicates that RUMAD effectively harnesses the increased diversity and collective intelligence of larger groups. While the token cost naturally increases with the number of agents, the framework remains efficient relative to the performance gains.}

\newtext{A more compelling insight emerges when scaling the agent pool under a fixed budget. By increasing the agent count from $N=6$ to $N=9$ while keeping the budget at $B=18$, we observe a simultaneous increase in MMLU accuracy (68.5\% $\to$ 70.3\%) and a significant decrease in token cost for both benchmarks (e.g., 22.5k $\to$ 17.6k on MMLU). This counter-intuitive result strongly demonstrates the adaptability of our RL controller. Given a larger pool of agents, the controller learns to allocate its fixed budget more judiciously, selecting a more specialized or diverse subset of agents to participate. This intelligent selection leads to a more efficient path to the solution, enhancing performance while reducing redundant communication.}

\newtext{Across experiments upon different swarm, RUMAD shows stable ability to find a more efficient and accurate policy without requiring extensive hyperparameter tuning. It suggests RUMAD learns a generalizable communication policy rather than overfitting to a specific setup, highlighting its hyperparameter insensitivity and flexibility for deployment in diverse environments.}

\section{Conclusion}

This paper introduces RUMAD, a principled reinforcement learning framework for adaptive topology control in multi-agent debate systems. By leveraging a content-agnostic observation design and a multi-objective reward scheme, RUMAD enables dynamic, efficient, and neutral management of agent interactions—successfully balancing accuracy, consensus, and computational efficiency under strict communication budgets. RUMAD also exhibits robust zero-shot generalization to diverse tasks and domains, showcasing strong transferability and practical deployment potential. Extensive experiments across MMLU, GSM8K, and GPQA demonstrate that RUMAD reduces token consumption by over 80\% compared to fully connected baselines, while maintaining or even improving collective reasoning accuracy against all baselines. 

These results establish RUMAD as a promising solution for scalable, cost-effective multi-agent reasoning, and open avenues for future research on principled \newtext{\textbf{coordination} and \textbf{communication mechanisms}} in large-scale \newtext{\textbf{multi-agent systems}}. \newtext{While RUMAD demonstrates strong performance, we acknowledge existing limitations. The centralized PPO controller, while efficient for typical MAD groups (6-8 agents), may face scalability challenges in scenarios with hundreds of agents. Future exploration about decentralized or hierarchical Multi-agent RL controllers is worth studying. }



\begin{acks}
This work was supported by Shenzhen Science and Technology Program (No. KJZD20240903100905008).

\end{acks}



\bibliographystyle{ACM-Reference-Format} 
\bibliography{sample}

@misc{hendrycks2020measuring,
  title={Measuring massive multitask language understanding},
  author={Hendrycks, Dan and Burns, Collin and Basart, Steven and Zou, Andy and Mazeika, Mantas and Song, Dawn and Steinhardt, Jacob},
  journal={arXiv preprint arXiv:2009.03300},
  year={2020}
}

@misc{cobbe2021training,
  title={Training verifiers to solve math word problems},
  author={Cobbe, Karl and Kosaraju, Vineet and Bavarian, Mohammad and Chen, Mark and Jun, Heewoo and Kaiser, Lukasz and Plappert, Matthias and Tworek, Jerry and Hilton, Jacob and Nakano, Reiichiro and others},
  journal={arXiv preprint arXiv:2110.14168},
  year={2021}
}

@misc{zeng2025s,
  title={S$^2$-{MAD}: {B}reaking the Token Barrier to Enhance Multi-Agent Debate Efficiency},
  author={Zeng, Yuting and Huang, Weizhe and Jiang, Lei and Liu, Tongxuan and Jin, Xitai and Tiana, Chen Tianying and Li, Jing and Xu, Xiaohua},
  journal={arXiv preprint arXiv:2502.04790},
  year={2025}
}

@misc{sun2023corex,
  title={Corex: {P}ushing the boundaries of complex reasoning through multi-model collaboration},
  author={Sun, Qiushi and Yin, Zhangyue and Li, Xiang and Wu, Zhiyong and Qiu, Xipeng and Kong, Lingpeng},
  journal={arXiv preprint arXiv:2310.00280},
  year={2023}
}

@misc{li2024improving,
  title={Improving Multi-Agent Debate with Sparse Communication Topology},
  author={Li, Yunxuan and Du, Yibing and Zhang, Jiageng and Hou, Le and Grabowski, Peter and Li, Yeqing and Ie, Eugene},
  journal={arXiv preprint arXiv:2406.11776},
  year={2024}
}

@misc{zhao2023survey,
  title={A survey of large language models},
  author={Zhao, Wayne Xin and Zhou, Kun and Li, Junyi and Tang, Tianyi and Wang, Xiaolei and Hou, Yupeng and Min, Yingqian and Zhang, Beichen and Zhang, Junjie and Dong, Zican and others},
  journal={arXiv preprint arXiv:2303.18223},
  year={2023}
}

@misc{naveed2023comprehensive,
  title={A comprehensive overview of large language models},
  author={Naveed, Humza and Khan, Asad Ullah and Qiu, Shi and Saqib, Muhammad and Anwar, Saeed and Usman, Muhammad and Akhtar, Naveed and Barnes, Nick and Mian, Ajmal},
  journal={arXiv preprint arXiv:2307.06435},
  year={2023}
}

@misc{achiam2023gpt,
  title={{GPT}-4 technical report},
  author={Achiam, Josh and Adler, Steven and Agarwal, Sandhini and Ahmad, Lama and Akkaya, Ilge and Aleman, Florencia Leoni and Almeida, Diogo and Altenschmidt, Janko and Altman, Sam and Anadkat, Shyamal and others},
  journal={arXiv preprint arXiv:2303.08774},
  year={2023}
}

@misc{glm2024chatglm,
  title={Chatglm: A family of large language models from glm-130b to glm-4 all tools},
  author={GLM, Team and Zeng, Aohan and Xu, Bin and Wang, Bowen and Zhang, Chenhui and Yin, Da and Zhang, Dan and Rojas, Diego and Feng, Guanyu and Zhao, Hanlin and others},
  journal={arXiv preprint arXiv:2406.12793},
  year={2024}
}

@misc{touvron2023llama,
  title={Llama 2: {O}pen foundation and fine-tuned chat models},
  author={Touvron, Hugo and Martin, Louis and Stone, Kevin and Albert, Peter and Almahairi, Amjad and Babaei, Yasmine and Bashlykov, Nikolay and Batra, Soumya and Bhargava, Prajjwal and Bhosale, Shruti and others},
  journal={arXiv preprint arXiv:2307.09288},
  year={2023}
}

@misc{guo2025deepseek,
  title={Deepseek-r1: Incentivizing reasoning capability in llms via reinforcement learning},
  author={Guo, Daya and Yang, Dejian and Zhang, Haowei and Song, Junxiao and Zhang, Ruoyu and Xu, Runxin and Zhu, Qihao and Ma, Shirong and Wang, Peiyi and Bi, Xiao and others},
  journal={arXiv preprint arXiv:2501.12948},
  year={2025}
}

@misc{jiang2024survey,
  title={A Survey on Large Language Models for Code Generation},
  author={Jiang, Juyong and Wang, Fan and Shen, Jiasi and Kim, Sungju and Kim, Sunghun},
  journal={arXiv preprint arXiv:2406.00515},
  year={2024}
}

@misc{liang2023encouraging,
  title={Encouraging divergent thinking in large language models through multi-agent debate},
  author={Liang, Tian and He, Zhiwei and Jiao, Wenxiang and Wang, Xing and Wang, Yan and Wang, Rui and Yang, Yujiu and Shi, Shuming and Tu, Zhaopeng},
  journal={arXiv preprint arXiv:2305.19118},
  year={2023}
}

@misc{hill2015real,
  title={Real conversations with artificial intelligence: A comparison between human--human online conversations and human--chatbot conversations},
  author={Hill, Jennifer and Ford, W Randolph and Farreras, Ingrid G},
  journal={Computers in human behavior},
  volume={49},
  pages={245--250},
  year={2015},
  publisher={Elsevier}
}

@misc{du2023improving,
  title={Improving factuality and reasoning in language models through multiagent debate},
  author={Du, Yilun and Li, Shuang and Torralba, Antonio and Tenenbaum, Joshua B and Mordatch, Igor},
  journal={arXiv preprint arXiv:2305.14325},
  year={2023}
}

@misc{liu2024groupdebate,
  title={Groupdebate: Enhancing the efficiency of multi-agent debate using group discussion},
  author={Liu, Tongxuan and Wang, Xingyu and Huang, Weizhe and Xu, Wenjiang and Zeng, Yuting and Jiang, Lei and Yang, Hailong and Li, Jing},
  journal={arXiv preprint arXiv:2409.14051},
  year={2024}
}

@misc{reinGPQAGraduateLevelGoogleProof2023,
  title = {{{GPQA}}: {{A Graduate-Level Google-Proof Q}}\&{{A Benchmark}}},
  shorttitle = {{{GPQA}}},
  author = {Rein, David and Hou, Betty Li and Stickland, Asa Cooper and Petty, Jackson and Pang, Richard Yuanzhe and Dirani, Julien and Michael, Julian and Bowman, Samuel R.},
  date = {2023-11-20},
  year = {2023},
  eprint = {2311.12022},
  eprinttype = {arXiv},
  eprintclass = {cs},
  journal={arXiv preprint arXiv:2311.12022},
  doi = {10.48550/arXiv.2311.12022},
  url = {http://arxiv.org/abs/2311.12022},
  urldate = {2025-08-01},
  pubstate = {prepublished}
}

@misc{schulmanProximalPolicyOptimization2017a,
  title = {Proximal {{Policy Optimization Algorithms}}},
  author = {Schulman, John and Wolski, Filip and Dhariwal, Prafulla and Radford, Alec and Klimov, Oleg},
  year = {2017},
  month = aug,
  number = {arXiv:1707.06347},
  eprint = {1707.06347},
  primaryclass = {cs},
  publisher = {arXiv},
  doi = {10.48550/arXiv.1707.06347},
  journal={arXiv preprint arXiv:1707.06347},
  archiveprefix = {arXiv}
}

@book{suttonReinforcementLearningIntroduction2018,
  title = {Reinforcement {{Learning}}: {{An Introduction}}},
  shorttitle = {Reinforcement {{Learning}}},
  author = {Sutton, Richard S. and Barto, Andrew G.},
  year = {2018},
  date = {2018-10},
  publisher = {A Bradford Book},
  location = {Cambridge, MA, USA},
  isbn = {978-0-262-03924-6},
  pagetotal = {552},
  address = {Cambridge, Massachusetts}
}

@misc{nussbaumNomicEmbedTraining2025,
  title = {Nomic {{Embed}}: {{Training}} a {{Reproducible Long Context Text Embedder}}},
  shorttitle = {Nomic {{Embed}}},
  author = {Nussbaum, Zach and Morris, John X. and Duderstadt, Brandon and Mulyar, Andriy},
  year = {2025},
  date = {2025-02-03},
  eprint = {2402.01613},
  eprinttype = {arXiv},
  eprintclass = {cs},
  journal={arXiv preprint arXiv:2402.01613},
  doi = {10.48550/arXiv.2402.01613},
  url = {http://arxiv.org/abs/2402.01613},
  urldate = {2025-05-20},
  pubstate = {prepublished}
}

@misc{schulmanHighDimensionalContinuousControl2018,
  title = {High-{{Dimensional Continuous Control Using Generalized Advantage Estimation}}},
  author = {Schulman, John and Moritz, Philipp and Levine, Sergey and Jordan, Michael and Abbeel, Pieter},
  date = {2018-10-20},
  year = {2018},
  eprint = {1506.02438},
  eprinttype = {arXiv},
  eprintclass = {cs},
  journal={arXiv preprint arXiv:1506.02438},
  doi = {10.48550/arXiv.1506.02438},
  url = {http://arxiv.org/abs/1506.02438},
  urldate = {2025-08-02},
  pubstate = {prepublished}
}

@misc{ahilan2019FeudalMultiAgentHierarchies,
  title = {Feudal {{Multi-Agent Hierarchies}} for {{Cooperative Reinforcement Learning}}},
  author = {Ahilan, Sanjeevan and Dayan, Peter},
  year = 2019,
  month = jan,
  number = {arXiv:1901.08492},
  eprint = {1901.08492},
  primaryclass = {cs},
  publisher = {arXiv},
  doi = {10.48550/arXiv.1901.08492},
  archiveprefix = {arXiv}
}

@misc{qian2025ScalingLargeLanguage,
  title = {Scaling {{Large Language Model-based Multi-Agent Collaboration}}},
  author = {Qian, Chen and Xie, Zihao and Wang, YiFei and Liu, Wei and Zhu, Kunlun and Xia, Hanchen and Dang, Yufan and Du, Zhuoyun and Chen, Weize and Yang, Cheng and Liu, Zhiyuan and Sun, Maosong},
  year = 2025,
  month = mar,
  number = {arXiv:2406.07155},
  eprint = {2406.07155},
  primaryclass = {cs},
  publisher = {arXiv},
  doi = {10.48550/arXiv.2406.07155},
  archiveprefix = {arXiv}
}

@misc{zhang2025AFlowAutomatingAgentic,
  title = {{{AFlow}}: {{Automating Agentic Workflow Generation}}},
  shorttitle = {{{AFlow}}},
  author = {Zhang, Jiayi and Xiang, Jinyu and Yu, Zhaoyang and Teng, Fengwei and Chen, Xionghui and Chen, Jiaqi and Zhuge, Mingchen and Cheng, Xin and Hong, Sirui and Wang, Jinlin and Zheng, Bingnan and Liu, Bang and Luo, Yuyu and Wu, Chenglin},
  year = 2025,
  month = apr,
  number = {arXiv:2410.10762},
  eprint = {2410.10762},
  primaryclass = {cs},
  publisher = {arXiv},
  doi = {10.48550/arXiv.2410.10762},
  archiveprefix = {arXiv}
}

@misc{zhang2025MultiagentArchitectureSearch,
  title = {Multi-Agent {{Architecture Search}} via {{Agentic Supernet}}},
  author = {Zhang, Guibin and Niu, Luyang and Fang, Junfeng and Wang, Kun and Bai, Lei and Wang, Xiang},
  year = 2025,
  month = jun,
  number = {arXiv:2502.04180},
  eprint = {2502.04180},
  primaryclass = {cs},
  publisher = {arXiv},
  doi = {10.48550/arXiv.2502.04180},
  archiveprefix = {arXiv}
}

\newpage
\onecolumn
\appendix
\definecolor{lightpurple}{rgb}{0.7, 0.6, 0.8}
\lstset{
    basicstyle=\ttfamily\small,
    breaklines=true,
    tabsize=2,
    showstringspaces=false,
    frame=single,
    rulecolor=\color{lightpurple},
}

\section{Detailed Hyper Parameters}
This section lists a detailed configuration of hyper parameters used in RUMAD framework.

\begin{table}[h]
\setlength{\tabcolsep}{5mm}
\centering
\caption{Weight parameter used in RUMAD multi-objective reward functions.}
\begin{tabular}{ccc}
\toprule
\textbf{Symbol} & \textbf{Description} & \textbf{Value} \\ \midrule
$\alpha_1$/$\beta_1$  & Accuracy              & 1.0/2.0     \\
$\alpha_2$/$\beta_2$  & Consensus             & 0.3/0.5     \\
$\alpha_3$  & Progress              & 0.1     \\
$\alpha_4$/$\beta_3$  & Efficiency            & 0.2/0.3     \\
$\alpha_5$/$\beta_4$  & Improvement           & 1.0/0.2      \\
$\alpha_6$  & Sparsity (penalty)    & 0.1      \\
\bottomrule           
\end{tabular}%
\label{tab:reward_weight_parameter}
\end{table}

\begin{table}[h]
\centering
\caption{Training hyperparameters used for the PPO-based topology controller of RUMAD.}
\begin{tabular}{ccc}
\toprule
{\color[HTML]{000000} \textbf{Symbol}} & \textbf{Description} & {\color[HTML]{000000} \textbf{Value}} \\ \midrule
$d_h$ & Hidden layer dimension & 128 \\
$\alpha$ & Learning rate & $1\times10^{-4}$ \\
$\gamma$ & Discount factor & 0.99 \\
$\lambda$ & GAE parameter & 0.95 \\
$\varepsilon$ & Policy clipping threshold & 0.20 \\
$c_1$ & Value loss coefficient & 0.01 \\
$c_2$ & Budget penalty coefficient & 0.01 \\
$E$ & Update epochs per rollout & 1\\
$b$ & Batch size per update & 10 \\
$L$ & Rollout length & 10 \\
\bottomrule
\end{tabular}%
\label{tab:hyperparameters}
\end{table}

\section{Detailed Debate Instance}

This section provides a detailed, step-by-step trace of a specific debate process for a multiple-choice question, which is visually summarized in Figure~\ref{fig:RUMAD_example} of the main text. The problem statement and the correct option are as follows.

\begin{lstlisting}[]
{
    "question": "To get to school in the morning, we ALL use this as our guide: A) Rand-McNally Road Atlas., B) world globe., C) road map., D) mental map..",
    "correct": "D"
}
\end{lstlisting}

\subsection{Independent Initialization}

In Round 0, all six agents ($\{0, 1, 2, 3, 4, 5\}$) execute the "Starting" prompt independently (see Table~\ref{tab:prompt}) to generate their initial responses to the problem. The generated responses from 3 different types of LLMs (see Section~\ref{sec:setup}) are shown below.

\begin{lstlisting}[]
{
    "Agent 0": "A question that gets to the heart of daily routines! To get to school in the morning, most people use a mental map (D) as their guide. A mental map is a cognitive representation of a person's knowledge about the layout of their surroundings, including the location of schools, homes, and other familiar places. It's a mental construct that is developed through repeated experiences and exposure to the environment. While options A, B, and C are all valid tools for navigating, they are not typically used as a guide for getting to school in the morning. Rand-McNally Road Atlas (A) is a physical book that provides detailed maps of roads and highways, but it's not something that people typically carry with them on a daily basis. A world globe (B) is a decorative item that shows the Earth's surface, but it's not a practical tool for getting to school. A road map (C) is a more portable version of a road atlas, but it's still not something that people typically use as a guide for getting to school. In contrast, a mental map is a flexible and dynamic representation of a person's knowledge about their surroundings. It's something that people develop through repeated experiences and exposure to the environment, and it's not something that can be easily replaced with a physical map or atlas. Therefore, the correct answer is: \\boxed{D) mental map.}",
    "Agent 1": "The answer is $\\boxed{C}$."
    "Agent 2": "The most accurate answer to the question of what is used as a guide to get to school in the morning is: D) mental map. A mental map is an individual's internal representation of their environment, which includes knowledge of locations, routes, and spatial relationships. When people navigate to school, they typically rely on their own mental maps rather than external tools like road maps or atlases. While a road map (C) could be used to plan a route, it is not the primary guide used during the actual navigation process. A world globe (B) is a three-dimensional representation of the Earth and is not practical for daily navigation. A Rand-McNally Road Atlas (A) is a physical map that could be used for planning, but it is not as commonly used for real-time navigation as a mental map. \\boxed{D}"
}
\end{lstlisting}

\subsection{Adaptive Sparse Debate}

The RUMAD framework proceeds through several debate rounds, where agents selectively update their responses based on the collective solutions shared from the previous round.

In Round 1, Agents 0, 1, 4, and 5 reuse their previous responses. Agents 2 and 3, however, generate new responses using the "Debate" prompt, incorporating and analyzing the solutions shared from Round 0. The generated responses are as follows.

\begin{lstlisting}[]
{
    "Agent 2": "To provide an updated answer, let's examine the responses from the num_agents step by step: 1. Agent 0 and Agent 3 both argue for the mental map (D) as the primary guide for getting to school in the morning. They emphasize that a mental map is a cognitive representation developed through repeated experiences and is not easily replaced by physical maps or atlases. 2. Agent 1 and Agent 4 suggest that the road map (C) is the correct answer. However, their arguments are not as detailed as those of Agents 0, 2, and 3, and they do not provide a clear explanation as to why a road map is the primary guide. 3. Agent 2 also supports the mental map (D) as the correct answer. They provide a similar rationale to Agents 0 and 3, emphasizing the individual's internal representation of the environment and the impracticality of using external tools like road maps or a world globe for daily navigation. Given the detailed and consistent support for the mental map (D) from Agents 0, 2, and 3, and the lack of a compelling argument from Agents 1 and 4, the most robust and persuasive answer is: \\boxed{D) mental map.}"
    "Agent 3": "After reviewing the responses from the other agents, I agree that the most accurate answer to the question of what is used as a guide to get to school in the morning is indeed: \\boxed{D) mental map.} The arguments presented by Agents 2, 3, and 5 are all consistent and well-reasoned, highlighting the importance of mental maps in daily navigation. They correctly point out that mental maps are internal representations of the environment, developed through repeated experiences and exposure, and are more practical and flexible than external tools like road maps or atlases. While Agents 1 and 5 did not provide detailed explanations, their answers still indicate that they believe the correct answer is D) mental map. In conclusion, based on the collective responses, I firmly believe that the most accurate answer is: \\boxed{D) mental map.}"
}
\end{lstlisting}

In Round 2, Only Agent 5 generates a new response, while all other agents reuse their previous solutions. The response of agent 5 is as follows.

\begin{lstlisting}
{
    "Agent 5": "After examining the responses from Agents 0, 2, 3, 4, and 5, it is clear that there is a strong consensus in favor of the mental map (D) as the primary guide for getting to school in the morning. Here's a step-by-step analysis of the arguments: - Agent 0 provides a clear explanation of what a mental map is and why it is the most practical guide for daily navigation, such as going to school. - Agent 2 summarizes the arguments from Agents 0, 2, and 3, reinforcing the idea that the mental map is the most persuasive answer due to its individual development and practicality. - Agent 4 offers a different perspective, suggesting that the road map (C) is the correct answer. However, this argument is not as detailed or persuasive as the ones presented by Agents 0, 2, and 5. - Agent 5 provides a comprehensive analysis, agreeing with Agents 0, 2, and 3 that the mental map (D) is the most accurate answer. They also explain why the other options (A, B, and C) are not as suitable for daily navigation. Given the detailed and consistent support for the mental map (D) from Agents 0, 2, 3, and 5, and the lack of a compelling argument from Agent 4, the most robust and persuasive answer is: \\boxed{D) mental map.}"
}
\end{lstlisting}

In Round 3, Agents 2, 3, 4, and 5 reuse their previous solutions, while Agents 0 and 1 generate new responses based on the updated collective solutions, which are as follows.

\begin{lstlisting}
{
    "Agent 0": "After examining the responses from the num_agents step by step, I agree with the consensus that the most accurate answer is: \\boxed{D) mental map.} The arguments presented by Agents 0, 2, 3, and 5 are all consistent and well-reasoned, highlighting the importance of mental maps in daily navigation. They correctly point out that mental maps are internal representations of the environment, developed through repeated experiences and exposure, and are more practical and flexible than external tools like road maps or atlases. While Agents 1 and 4 did not provide detailed explanations, their answers still indicate that they believe the correct answer is D) mental map. In conclusion, based on the collective responses, I firmly believe that the most accurate answer is: \\boxed{D) mental map.}",
    "Agent 1": "After examining the responses from the other agents, I agree that the most accurate answer to the question of what is used as a guide to get to school in the morning is indeed: The answer is $\\boxed{D}$."
}
\end{lstlisting}

In Round 4, all six agents reuse their previous responses, leading to no new responses being generated. 

In Round 5, only Agents 0 and 2 generate updated responses which are as follows, further reinforcing the consensus reached in the previous rounds.

\begin{lstlisting}
{
    "Agent 0": "After examining the responses from the other agents, I agree that the most accurate answer to the question of what is used as a guide to get to school in the morning is indeed: The arguments presented by Agents 0, 2, 3, and 5 are all consistent and well-reasoned, highlighting the importance of mental maps in daily navigation. They correctly point out that mental maps are internal representations of the environment, developed through repeated experiences and exposure, and are more practical and flexible than external tools like road maps or atlases. While Agents 1 and 4 did not provide detailed explanations, their answers still indicate that they believe the correct answer is D) mental map. In conclusion, based on the collective responses, I firmly believe that the most accurate answer is: \\boxed{D) mental map.}",
    "Agent 2": "After carefully examining the responses from Agents 0, 2, 3, 4, and 5, it is evident that there is a strong consensus supporting the mental map (D) as the primary guide for daily navigation, such as getting to school in the morning. - Agent 0 provides a comprehensive explanation of the concept of a mental map, emphasizing its role as an internal representation of the environment developed through repeated experiences. This internal guide is deemed more practical and flexible than external tools like road maps or atlases. - Agent 2 reinforces the argument for the mental map by summarizing the points made by Agents 0, 2, and 3, further solidifying the case for its practicality and individual development. - Agent 3 adds to the argument by highlighting the cognitive aspect of mental maps, which are not easily replaced by physical maps or atlases, making them ideal for daily navigation. - Agent 4 presents a contrasting view, suggesting that the road map (C) is the correct answer. However, this argument lacks the depth and detail of the mental map explanation provided by Agents 0, 2, 3, and 5. - Agent 5 offers a comprehensive analysis, agreeing with the mental map argument and explaining why the other options are less suitable for daily navigation. Given the detailed and consistent support for the mental map (D) from Agents 0, 2, 3, and 5, and the lack of a compelling argument from Agent 4, the most robust and persuasive answer is: \\boxed{D) mental map.}"
}
\end{lstlisting}

Following the completion of the adaptive debate rounds, the final aggregated conclusion, "D) mental map.", is determined via majority voting among all agents.

\section{Prompt Configuration}

This section details the system and stage-specific prompts used to configure the RUMAD framework for the experiments. Table~\ref{tab:output_format} outlines the specific output format requirements for the various datasets used, and Table~\ref{tab:prompt} lists the prompts utilized in each stage of the debate process.

\begin{table}[h]
    \centering
    \caption{Output format requirements across different tasks.}
    \begin{tabularx}{\columnwidth}{l X}
        \toprule
        \textbf{Task} & \textbf{Output Format Requirements} \\
        \midrule     
        MMLU \& GPQA & Put your final choice in the form \textbackslash boxed\{\{answer\}\} at the end of your response. \\
        \midrule
        GSM8K & Your final answer should be a single numerical number, in the Form \textbackslash boxed\{\{answer\}\}, at the end of your response. \\
        \bottomrule
    \end{tabularx}
    \label{tab:output_format}
\end{table}

\begin{table*}[h]
    \centering
    \caption{Prompts used in each stage of the debate.}
    \begin{tabularx}{\textwidth}{l l X}
        \toprule
        \textbf{Type} & \textbf{Task} & \textbf{Prompt} \\
        \midrule
        System & All & Welcome to the debate! You are a seasoned debater with expertise in succinctly and persuasively expressing your viewpoints. You will be assigned to debate groups, where you will engage in discussions with fellow participants. The outcomes of each group's deliberations will be shared among all members. It is crucial for you to leverage this information effectively in order to critically analyze the question at hand and ultimately arrive at the correct answer. Best of luck! \\
        \midrule
        Starting & All & Can you answer the following question as accurately as possible? $<$Question$>$ Explain your answer. $<$Output Format$>$. \\
        \midrule     
        Debate & All & These are the solutions from other agents: $<$other agent responses$>$. Based on the above responses with their indicated importance, can you provide an updated answer? Examine all solutions step by step. $<$Output Format$>$. \\
        \bottomrule
    \end{tabularx}
    \label{tab:prompt}
\end{table*}

\end{document}